\documentclass[10pt, a4paper]{article}

\usepackage[final]{lrec2026} 

\usepackage{booktabs}

\usepackage{pifont}    
\usepackage{arydshln}   
\usepackage{array}     
\usepackage{multirow}   
\usepackage{enumitem}   

\usepackage[most, listings, breakable]{tcolorbox}

\newcommand{\citelr}{\citelanguageresource}

\title{Multilingual KokoroChat: A Multi-LLM Ensemble Translation Method for Creating a Multilingual Counseling Dialogue Dataset}

\name{Ryoma Suzuki, Zhiyang Qi, Michimasa Inaba} 

\address{The University of Electro-Communications \\
     1-5-1, Chofugaoka, Chofu, Tokyo, Japan \\
     \{r-suzuki, qizhiyang, m-inaba\}@uec.ac.jp\\}

\abstract{
To address the critical scarcity of high-quality, publicly available counseling dialogue datasets, we created Multilingual KokoroChat by translating KokoroChat, a large-scale manually authored Japanese counseling corpus, into both English and Chinese.
A key challenge in this process is that the optimal model for translation varies by input, making it impossible for any single model to consistently guarantee the highest quality. In a sensitive domain like counseling, where the highest possible translation fidelity is essential, relying on a single LLM is therefore insufficient. To overcome this challenge, we developed and employed a novel multi-LLM ensemble method. Our approach first generates diverse hypotheses from multiple distinct LLMs. A single LLM then produces a high-quality translation based on an analysis of the respective strengths and weaknesses of all presented hypotheses.
The quality of ``Multilingual KokoroChat'' was rigorously validated through human preference studies. These evaluations confirmed that the translations produced by our ensemble method were preferred from any individual state-of-the-art LLM. This strong preference confirms the superior quality of our method's outputs. The Multilingual KokoroChat is available at \url{https://github.com/UEC-InabaLab/MultilingualKokoroChat}.
 \\ \newline  }

\begin{document}

\maketitleabstract

\section{Introduction}
The COVID-19 pandemic has exacerbated the global mental health crisis, worsening numerous factors that contribute to psychological distress~\cite{santomauro2021global}. 
Despite the growing need, access to professional mental healthcare remains a significant challenge for many, largely due to a shortage of skilled counselors. 
To bridge this gap, mental health support systems leveraging Large Language Models (LLMs) have emerged as a promising solution, showing great potential in alleviating distress and helping individuals navigate personal challenges~\cite{sharma2023human}. 
However, the development of such systems is severely hampered by the scarcity of publicly available, high-quality, human-authored counseling dialogue datasets. To address this limitation, our work aims to extend the large-scale Japanese counseling dialogue corpus, KokoroChat~\citelr{qi-etal-2025-kokorochat}, into a multilingual resource through translation.

For this multilingual extension, machine translation using LLMs is a powerful tool \cite{enis2024llmnmtadvancinglowresource,he2024exploring,xuparadigm}. 
However, the optimal LLM for generating the best translation is inconsistent across different inputs, as models exhibit varying strengths in handling specific contexts and expressions \cite{jiang-etal-2023-llm,wang2024mixtureofagentsenhanceslargelanguage}. 
Consequently, even a single state-of-the-art model cannot always guarantee the highest translation quality. This issue becomes particularly critical in sensitive domains like counseling, where superior quality and stability are paramount to avoid the risk of misleading translations. 
To overcome this challenge, we propose a novel multi-LLM ensemble method. 
Our approach feeds the source text, along with translations from multiple LLMs, into a final LLM that synthesizes their strengths and mitigates their weaknesses to generate a more refined output. 
Using this method, we have developed the first multilingual psychological counseling dataset, named \textbf{Multilingual KokoroChat}, which provides a new foundational resource for cross-lingual counseling research and the development of multilingual dialogue systems.

\section{Related Work}

\subsection{Psychological Counseling Datasets}
\begin{table*}[t!]
\centering
\small
\renewcommand{\arraystretch}{1.2}
\setlength{\tabcolsep}{5pt}
\begin{tabular}{lccccc} 
\toprule
\textbf{Dataset} & \textbf{Method} & \textbf{Open-source} & \textbf{Language} & \textbf{Dialogues} & \textbf{Avg. utterances} \\
\midrule
HealMe \cite{xiao-etal-2024-healme}    & LLM & \ding{51} & En & 1,300 & 6.0 \\ 
ESD-CoT \cite{zhang-etal-2024-escot}    & LLM & \ding{51} & En & 1,708 & 23.4 \\ 
C\textsc{actus} \cite{lee-etal-2024-cactus} & LLM & \ding{51} & En & 31,577 & 31.5 \\ 
SMILECHAT \cite{qiu-etal-2024-smile}    & LLM & \ding{51} & Zh & 55,165 & 33.2 \\ 
A\textsc{ug}ESC \cite{zheng-etal-2023-augesc} & LLM & \ding{51} & En & 65,000 & 26.7 \\ 
\noalign{\vskip 3pt}
\cdashline{1-6}
\noalign{\vskip 3pt}
Anno-MI \cite{AnnoMi}           & Human & \ding{51} & En & 133  & 72.9 \\ 
ESConv \cite{liu-etal-2021-towards}    & Human & \ding{51} & En & 1,300 & 29.5 \\ 
Client-Reactions \cite{li-etal-2023-understanding} & Human & \ding{55} & Zh & 2,382 & 78.5 \\ 
\textbf{KokoroChat} \cite{qi-etal-2025-kokorochat} & Human & \ding{51} & Ja & \textbf{6,589} & \textbf{91.2} \\ 
\noalign{\vskip 3pt}
\cdashline{1-6}
\noalign{\vskip 3pt}
PsyDial \cite{qiu-lan-2025-psydial}    & Hybrid & \ding{51} & Zh & 2,382 & 75.6 \\ 
\multirow{2}{*}{\textbf{Multilingual KokoroChat}} & \multirow{2}{*}{Hybrid} & \multirow{2}{*}{\ding{51}} & \textbf{Zh/Ja} & \textbf{6,565} & \textbf{91.2} \\ 
                                                  &                         &                            & \textbf{En/Ja} & \textbf{6,582} & \textbf{91.2} \\
\bottomrule
\end{tabular}
\caption{\label{tab:dataset_comparison}
Comparison of psychological counseling datasets categorized by construction method: LLM-based (top), human-collected (middle), and hybrid (bottom). 
We further build and release a multilingual version based on the high-quality human-collected dataset \textit{KokoroChat}. Note that the slight reduction in the number of dialogues for the multilingual versions is due to the exclusion of content that triggered the LLM's safety filters. }
\end{table*}

In recent years, psychological counseling has attracted increasing attention in the field of Natural Language Processing (NLP). Early studies mainly focused on empathic response generation, where systems aim to produce emotionally appropriate responses by recognizing the user's affective state \cite{rashkin-etal-2019-towards, sharma-etal-2020-computational, zheng-etal-2021-comae}. With the rapid advancement of LLMs, research interest has gradually shifted from simple empathy generation to more complex modeling of counseling dialogues. Consequently, a number of LLM-based counseling chat systems have been proposed, such as ChatCounselor \cite{ChatCounselor}, MeChat \citelr{qiu-etal-2024-smile}, and SoulChat \citelr{chen-etal-2023-soulchat}.

The performance of these systems largely depends on the availability of high-quality counseling datasets, making data construction a central research issue. Existing efforts for building such datasets can be broadly categorized into two approaches: \textbf{human-created} and \textbf{LLM-augmented}.

Human-created datasets are typically collected and curated from real human dialogues. For instance, Anno-MI \citelr{AnnoMi} was constructed from motivational interviewing conversations extracted from online videos; ESConv \citelr{liu-etal-2021-towards} collected emotional support dialogues from crowdworkers who had undergone professional skill training; ClientReactions \citelr{li-etal-2023-understanding} was derived from real counselor–client interactions on online counseling platforms; and KokoroChat \citelr{qi-etal-2025-kokorochat} was systematically collected through role-playing by trained counselors.

On the other hand, many studies adopt LLM-based automatic generation, where the model simultaneously plays both counselor and client to create data efficiently and at scale. Representative examples include HealMe \citelr{xiao-etal-2024-healme} and C\scalebox{0.8}{ACTUS} \citelr{lee-etal-2024-cactus}, which generate multi-turn dialogues using prompts inspired by Cognitive Behavioral Therapy (CBT); ESD-CoT \citelr{zhang-etal-2024-escot}, which generates full conversations based on scenarios extracted from existing datasets; SMILECHAT \citelr{qiu-etal-2024-smile}, which expands single-turn Q\&A pairs into multi-turn dialogues; and A\scalebox{0.8}{UG}ESC \citelr{zheng-etal-2023-augesc}, which formulates data augmentation as a dialogue completion task.
In addition, PsyDial \citelr{qiu-lan-2025-psydial} adopts a \textbf{hybrid construction approach} that regenerates client utterances using LLMs based on ClientReactions, producing a semi-authentic dataset that preserves realism while protecting privacy.

As shown in Table~\ref{tab:dataset_comparison}, although automatically generated data are advantageous in scalability and efficiency, they remain limited in authenticity—often with fewer dialogue turns—and in content diversity \cite{zheng-etal-2024-self}. Therefore, this study focuses on human-collected dialogues and adopts KokoroChat \citelr{qi-etal-2025-kokorochat}, the largest and most professionally curated open counseling dataset, as the foundation. 

\subsection{LLM Ensemble}
Ensemble learning, a method that combines multiple models to surpass the performance of any single model, has long been recognized as an effective technique in machine learning \cite{LITTLESTONE1994212, GANAIE2022105151} and has been shown to improve system performance in machine translation \cite{zhou-etal-2017-neural}.

LLM ensembles initially emerged through selection-based approaches, such as the PairRanker module in LLM-Blender \cite{jiang-etal-2023-llm} and MBR (Minimum Bayes Risk) decoding \cite{fernandes-etal-2022-quality, farinhas-etal-2023-empirical}, which choose the best output from multiple candidate outputs generated by multiple LLMs. However, this approach has a fundamental limitation: it cannot produce output quality that surpasses that of the best initial candidate. Furthermore, for our task of translating entire dialogues, selection-based approaches are inadequate. Because the dialogue is assembled from outputs of multiple models, terminology and stylistic tone can become inconsistent across turns.

\begin{figure}[t]
 \centering
 \includegraphics[width=7.5cm]{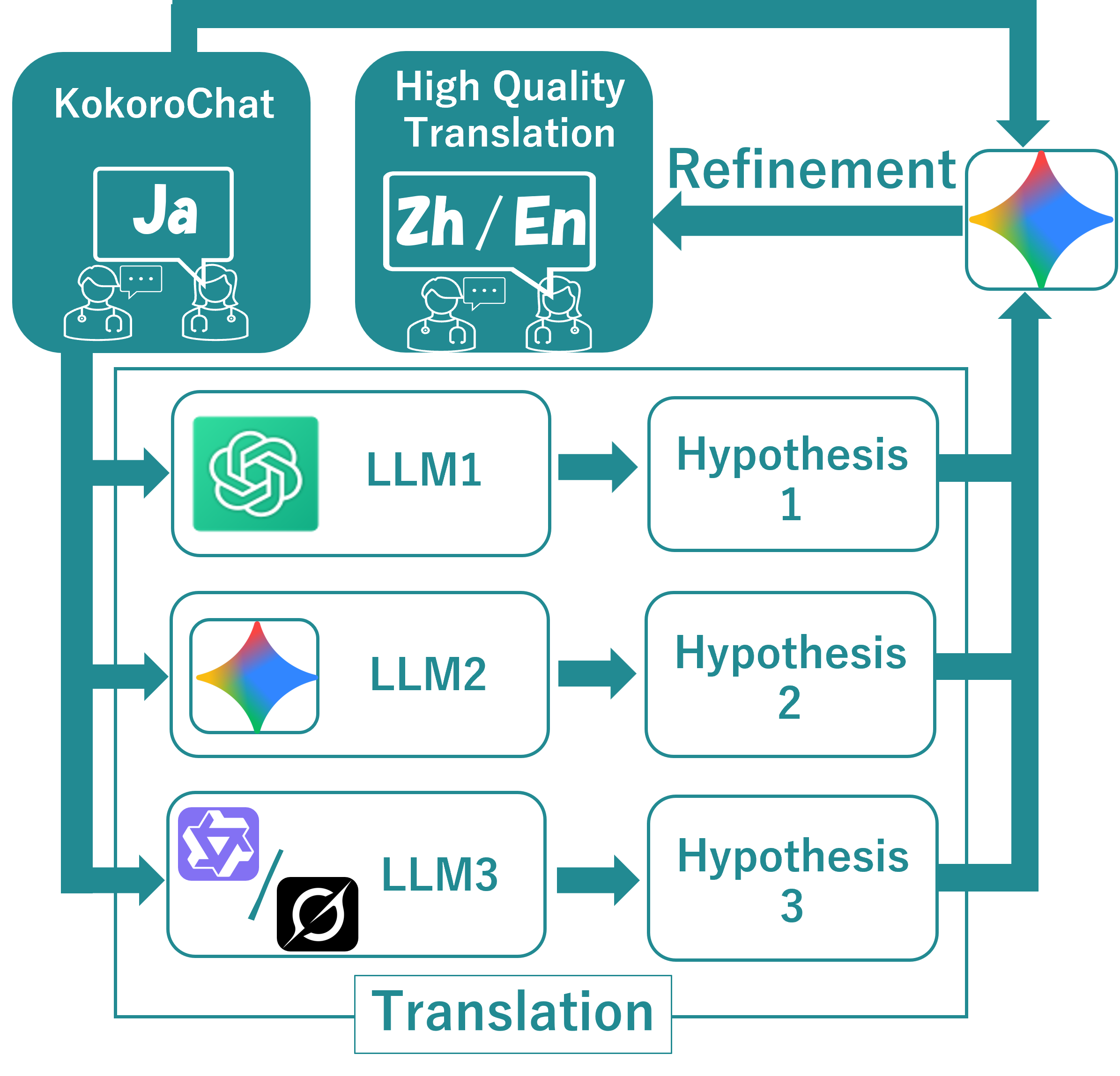}
 \caption{Proposed Multi-LLM Ensemble Translation Method}
 \label{flow chart}
\end{figure}

To overcome these limitations, research has progressed toward fusion-based approaches that integrate the strengths of multiple candidates to synthesize a new output superior to any individual candidate. For example, the GenFuser module in LLM-Blender fuses top-ranked candidates to generate a refined output \cite{jiang-etal-2023-llm}, and a layered Mixture-of-Agents (MoA), in which agents use prior-layer outputs as auxiliary context, enables hierarchical collaboration \cite{wang2024mixtureofagentsenhanceslargelanguage}; both approaches consistently outperform single models.
EVA (Ensemble via Vocabulary Alignment), which addresses lexical inconsistencies between different LLMs by fusing probability distributions at the token level, has demonstrated superior performance over selection-based ensemble methods like MBR and PairRanker, demonstrating the advantages of fusion-based approaches \cite{xu-etal-2024-bridging}.

Furthermore, in Medical QA tasks, LLM-Synergy surpassed the performance of individual models by applying a boosting-based weighted majority vote \cite{info:doi/10.2196/70080}.
The vote aggregated outputs from LLMs that were dynamically selected based on question context through clustering.
These successes in specialized domains strongly support applying a tailored ensemble method to counseling dialogue translation.

Our method is a fusion-based approach, but it is distinguished by its direct, analysis-driven refinement process. Specifically, the refiner LLM is instructed to analyze and articulate both the strengths and weaknesses of all presented translation candidates. Based on this generated analysis, the final translation is synthesized. Our method is distinguished from existing research by incorporating refiner LLMs' advanced reasoning capabilities directly into the fusion process itself.

\section{KokoroChat}
The foundation of our work is KokoroChat \citelr{qi-etal-2025-kokorochat}, a large-scale, publicly available collection of manually authored Japanese text-based counseling dialogues created through role-playing. 
The participants are professional counselors and trainee counsellors. 
The corpus comprises 6,589 dialogue sessions, each lasting 60 minutes, totalling approximately 600,000 utterances.
The corpus covers a wide range of counseling topics, including issues related to school, work, family relationships, romance, and financial problems. 
A key feature of KokoroChat is that every dialogue is annotated with 20-item feedback scores provided by the participant playing the client role. 
These scores enable researchers to develop counseling dialogue systems using only high-quality data or automatic evaluation models for counselling dialogues. 
Given its status as the world's largest manually curated counseling corpus and its unique integration of feedback scores, KokoroChat is a critical resource. 
We argue that extending this corpus into a multilingual format will provide significant value to the international research community.

\begin{table}[t]
\centering
\footnotesize
\begin{tabular}{p{0.95\linewidth}}
\toprule
\textbf{\# Data Description} \\
-- This is Japanese text counseling data from role-playing sessions where counselors acted as both counselor and client. \\
-- Each line is separated by a colon (':'). The left side indicates the role name, and the right side is the utterance. \\ \relax
\\
\textbf{\# Translation Instructions} \\
-- As a professional translator, translate this data into \textbf{\{target language\}}. \\
-- For the translation, please use polite and natural expressions that are appropriate for a counseling context. \\
\bottomrule
\end{tabular}
\caption{
Prompt for Hypothesis Generation
}
\label{fig:instruction-hypothesis}
\end{table}

\section{Multi-LLM Ensemble Translation}

To construct a high-quality multilingual counseling dialogue dataset, this study aims to address the challenge that the optimal LLM varies depending on the input, meaning no single model can consistently guarantee the best possible quality. 
Even high-performance models have distinct strengths and weaknesses, leading to inconsistent output quality. This instability is particularly unacceptable in sensitive domains like counseling, where the risk of mistranslation must be minimized.

To overcome this challenge, we propose a novel multi-LLM ensemble method, as shown in Figure \ref{flow chart}. 
The approach consists of two key stages: generating diverse translation candidates (hypotheses) and refining them through integration.

\begin{table}[t]
\centering
\footnotesize
\begin{tabular}{p{0.95\linewidth}}
\toprule
\textbf{\# Data Description} \\
-- This data includes Japanese text counseling data and its \{\textbf{target language}\} translation candidates. \\
-- Japanese text counseling data was collected through role-playing between counselors acting as counselor and client. \\ \relax
\\
\textbf{\# Input Data Format} \\
-- The input is a list of dictionary objects. \\
\textbf{\#\# Keys of each dictionary} \\
\hspace{1em}-- 'role': Role of the speaker ('Counselor' or 'Client') \\
\hspace{1em}-- 'source': Japanese original text \\
\hspace{1em}-- 'hypothesis1', 'hypothesis2', 'hypothesis3': \textbf{\{target language\}} translation candidates \\ \relax
\\
\textbf{\# Evaluation Instructions} \\
-- You are a professional translator, evaluate the \textbf{\{target language\}} translation candidates. \\
-- For each utterance, follow these steps for evaluation: \\
\hspace{1em}\textbf{1. Analysis of Each Translation Candidate} \\
\hspace{1em}-- Compare each translation candidate and describe specifically which parts are superior. \\
\hspace{1em}-- Describe specifically which parts need improvement. \\
\hspace{1em}\textbf{2. Construction of an Improved Translation} \\
\hspace{1em}-- Based on your analysis, synthesize a revised translation by combining the strengths of both candidates. \\
\hspace{1em}-- Make corrections based on the areas for improvement you identified. \\
\hspace{1em}-- Ensure consistent terminology to maintain consistency throughout the translation. \\
\bottomrule
\end{tabular}
\caption{
Prompt for Integration and Refinement
}
\label{fig:instruction-refinement}
\end{table}

\subsection{Diverse Hypothesis Generation}
The objective of this stage is not merely to obtain multiple translations, but to construct a diverse and high-quality set of translation hypotheses.
To achieve this, we instruct three distinct LLMs, using the system prompt detailed in Table \ref{fig:instruction-hypothesis}, to generate three independent translation hypotheses per language.

Crucially, we process each entire dialogue as a single input rather than translating utterance by utterance. This approach is essential for preserving dialogue-level phenomena often lost in sentence-by-sentence translation. Providing the complete conversational history enables the LLMs to generate contextually coherent translations, maintain consistent terminology, and preserve the distinct linguistic styles of each speaker. The result is a translation that functions as a single, cohesive, and natural-flowing conversation, not merely a collection of disjointed sentences.

\subsection{Integration and Refinement}

In the second stage, a specialized ``refiner'' LLM is tasked with integration and refinement. The objective of this stage is not simply to select the best among the three candidates, but to create a new translation that surpasses any of the original hypotheses by synthesizing their respective strengths.

To achieve this, the refiner LLM is provided with both the original Japanese dialogue and the three translation hypotheses generated in the first stage. Crucially, these inputs are processed at the dialogue level to ensure the refiner LLM fully comprehends the overall context, including the flow of conversation and the consistency of speaker roles.

While maintaining this global context, the model is prompted to perform its analysis and synthesis on an utterance-by-utterance basis. Specifically, following the system prompt detailed in Table \ref{fig:instruction-refinement}, we instruct the model to first analyze and articulate the strengths and weaknesses of each of the three candidate translations for a given utterance. Based on this analysis, it is then directed to synthesize a final, polished translation for that utterance, integrating the identified strengths while mitigating the noted shortcomings.

\section{Experiments}

To validate proposed multi-LLM ensemble method and assess the quality of the resulting Multilingual KokoroChat, we conducted experiments on translation into English and Chinese. 
We compared the output of our method with that of single LLMs using both automatic and human evaluation. 
The automatic evaluation provides a large-scale, objective measure of quality, while the human evaluation captures crucial nuances such as naturalness and contextual appropriateness, which automated metrics often fail to capture.

\subsection{Experimental Settings}

For the English translation, we selected three models representing the state-of-the-art at the time of manuscript preparation: GPT-5 (gpt-5-2025-0907) \footnote{https://platform.openai.com/docs/models/gpt-5}, Gemini 2.5 Pro (gemini-2.5-pro) \footnote{https://ai.google.dev/gemini-api/docs/models} and Grok-4 (grok-4-0709) \footnote{https://docs.x.ai/docs/models}. For Chinese translation, we replaced Grok-4 with Qwen-Plus (qwen-plus-2025-07-28) \footnote{https://www.alibabacloud.com/help/ja/model-studio/what-is-qwen-llm}, which demonstrates superior performance and produces more stable outputs for Chinese-language tasks. 
This approach allows us to collect a set of hypotheses for each language that ensures both quality and diversity.

We selected Gemini 2.5 Pro as the refiner LLM, tasked with integrating and polishing the candidate translations. 
This decision was based on a preliminary analysis where Gemini 2.5 Pro demonstrated a superior ability over other models in comparing multiple texts, identifying specific areas for improvement, and synthesizing these insights into a cohesive and improved output.
All outputs were generated using the default parameter settings for each respective model. 

\begin{table}[t]
\centering
\begin{tabular}{lcc}
\toprule
\textbf{System} & \textbf{MetricX↓} & \textbf{XCOMET↑} \\
\midrule
GPT  & 3.409$^*$ & 0.921$^*$    \\
Gemini & 3.356$^*$ & \textbf{0.927}$^*$ \\
Grok  & 3.369$^*$ & \textbf{0.927}  \\
\midrule
Proposed & \textbf{3.345} & 0.926      \\
\bottomrule
\end{tabular}
\caption{Automatic Evaluation Results for Japanese-to-English Translation. $^*$indicates a statistically significant performance difference $p < 0.01$ compared to the proposed method.}
\label{tab:stats_english}
\end{table}

\begin{table}[t]
\centering
\begin{tabular}{lcc}
\toprule
\textbf{System} & \textbf{MetricX↓} & \textbf{XCOMET↑} \\
\midrule
GPT  & 2.547$^*$ & 0.859$^*$    \\
Gemini & 2.483$^*$ & 0.870$^*$    \\
Qwen  & 2.476$^*$ & \textbf{0.871}  \\
\midrule
Proposed & \textbf{2.445} & \textbf{0.871}  \\
\bottomrule
\end{tabular}
\caption{Automatic Evaluation Results for Japanese-to-Chinese Translation. $^*$indicates a statistically significant performance difference $p < 0.01$ compared to the proposed method.}
\label{tab:stats_chinese}
\end{table}

\subsection{Automated Evaluation}

The source KokoroChat dataset lacks human translations. Consequently, conventional reference-based metrics such as BLEURT \cite{sellam-etal-2020-bleurt} or BARTScore \cite{NEURIPS2021_e4d2b6e6} were not applicable. Therefore, we employed two reference-free evaluation metrics: XCOMET-QE \footnote{https://huggingface.co/Unbabel/XCOMET-XL} \cite{guerreiro-etal-2024-xcomet} and MetricX-QE \footnote{https://huggingface.co/google/metricx-24-hybrid-xl-v2p6} \cite{juraska-etal-2024-metricx}. Both were chosen for their reported high correlation with human preferences in the WMT24 Metrics Shared Task \cite{freitag-etal-2024-llms}. XCOMET scores range from 0 to 1, where higher is better, while MetricX scores range from 0 to 25, where lower is better.

For this evaluation, we used 200 randomly sampled dialogues from KokoroChat. To remove the trivial samples that would not yield meaningful differences in evaluation scores, we filtered this set to include only utterances for which at least three of the four systems (proposed method and the three single LLMs) generated unique translations. The final evaluation datasets consisted of 13,106 English utterances and 14,544 Chinese utterances.

\begin{figure}[t]
 \centering
 \includegraphics[width=0.48\textwidth]{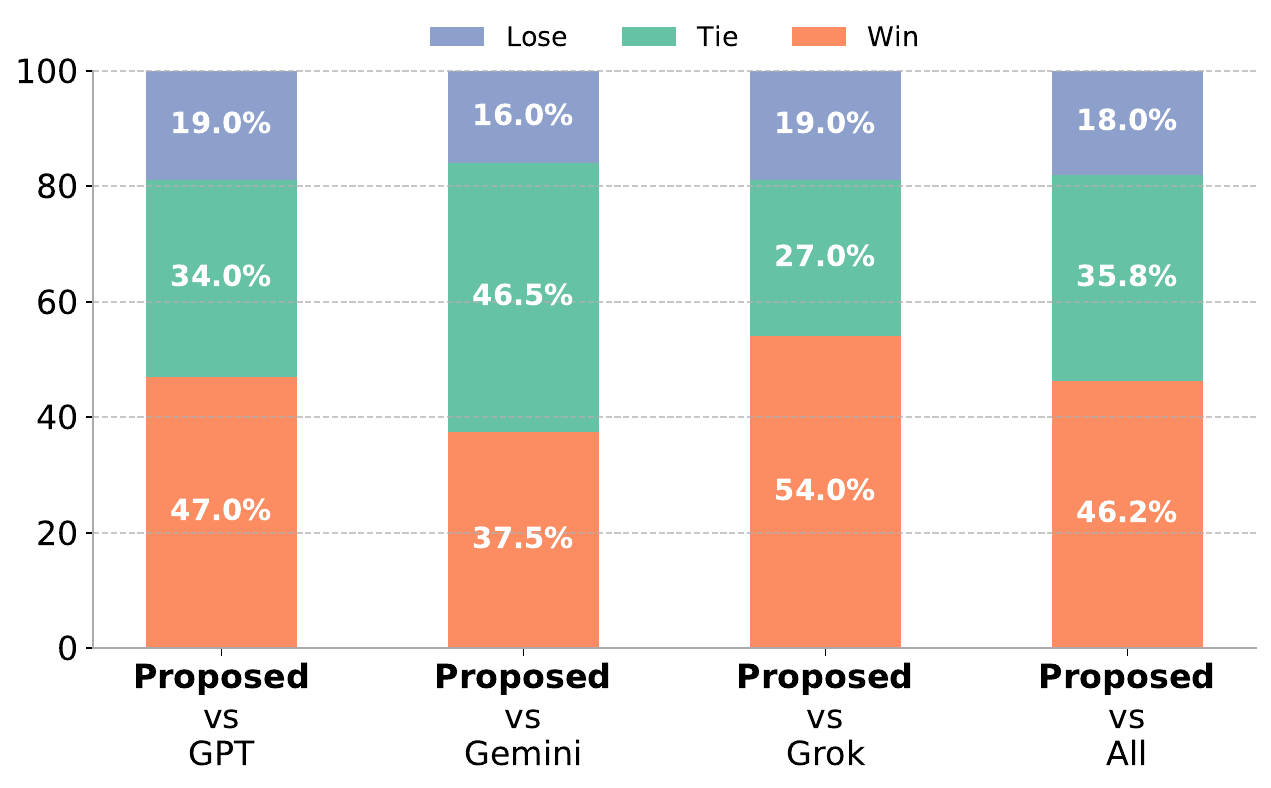}
 \caption{Human Evaluation Results for Japanese-to-English Translation} 
 \label{human_english}
\end{figure}

\begin{figure}[t]
 \centering
 \includegraphics[width=0.48\textwidth]{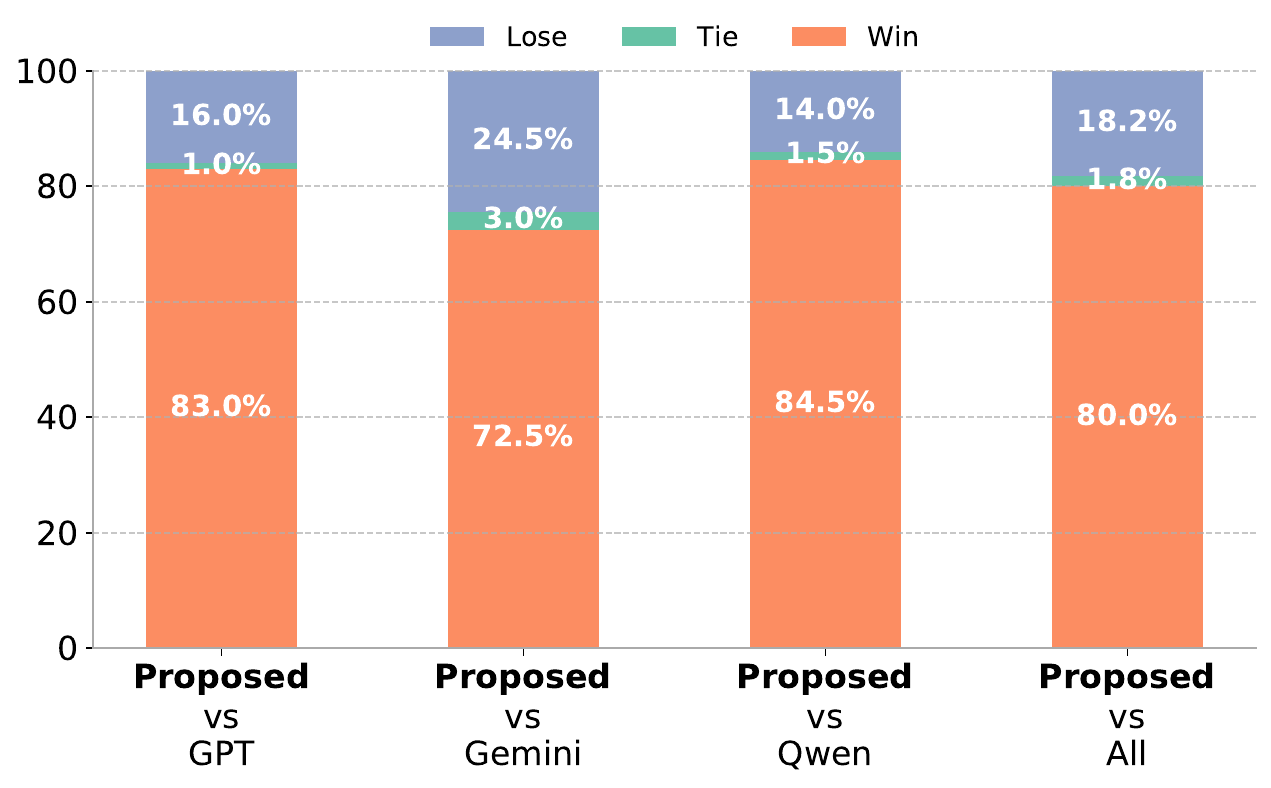}
 \caption{Human Evaluation Results for Japanese-to-Chinese Translation}
 \label{human_chinese}
\end{figure}

\begin{figure*}[t]
 \centering
 \includegraphics[width=0.95\textwidth]{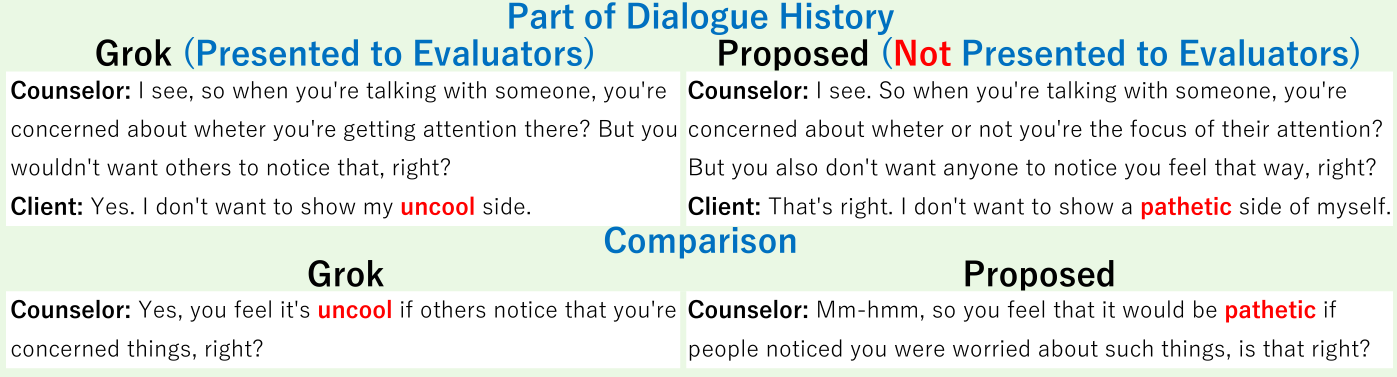}
 \caption{Comparison of Japanese-to-English translation where the proposed method was judged inferior to Grok. This judge resulted from a contextual mismatch inherent in the randomized experimental design: the use of "pathetic" for internal consistency conflicted with the term "uncool" in the context randomly assigned to the evaluators.}
 
 \label{case_bad_english}
\end{figure*}

\subsection{Results of Automated Evaluation}
The results are presented in Tables \ref{tab:stats_english} and \ref{tab:stats_chinese}.
To ensure statistical rigor, we performed a pairwise significance test using the Wilcoxon Signed-Rank Test \cite{PairWise} between our method and each single-LLM baseline. The Bonferroni correction \cite{bonferroni1936teoria} was applied to counteract the increased risk of Type I errors from multiple comparisons; specifically, all p-values were multiplied by three. 

Based on the MetricX, proposed method achieved the best (lowest) scores for both English and Chinese. For the Chinese translation, this superiority was statistically significant over all single-LLM baselines.

In the XCOMET metrics, our method tied with Qwen for the top performance in Chinese, achieving statistically significant improvements over both GPT and Gemini. For English, while marginally behind Gemini and Grok, our method still demonstrated a statistically significant improvement over GPT.

Collectively, these results provide strong quantitative evidence that the translations generated by proposed method are consistently superior to, or at worst comparable with, those from single-LLM approaches.

\begin{figure*}[t]
 \centering
 \includegraphics[width=0.95\textwidth]{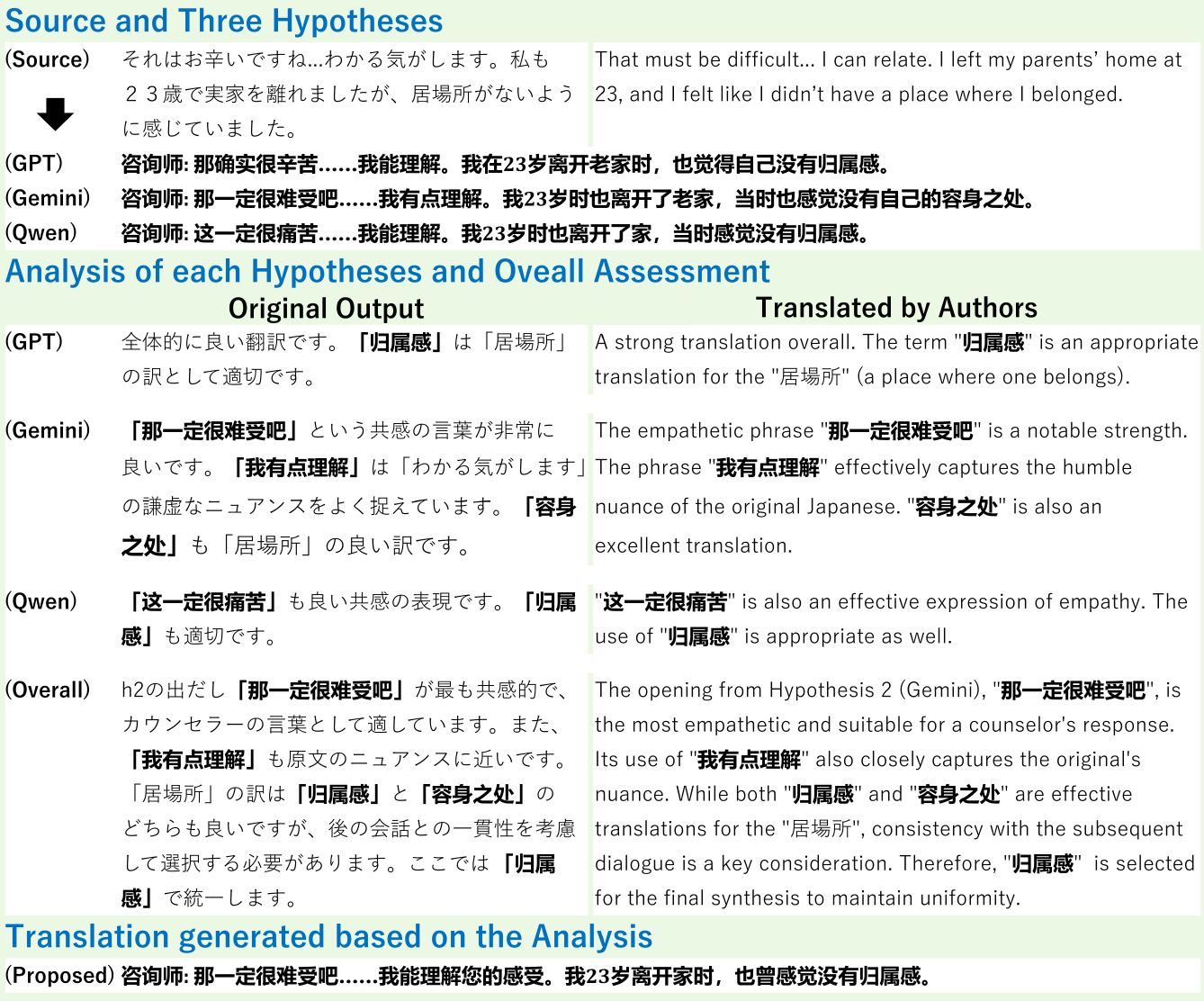}
 \caption{Japanese-to-Chinese translation example 1. This demonstrates the analysis and synthesis process that produced a final translation highly preferred by human evaluators over any of the three single-LLM hypotheses.}
 \label{case_good_chinese}
\end{figure*}

\begin{figure*}[t]
 \centering
 \includegraphics[width=0.95\textwidth]{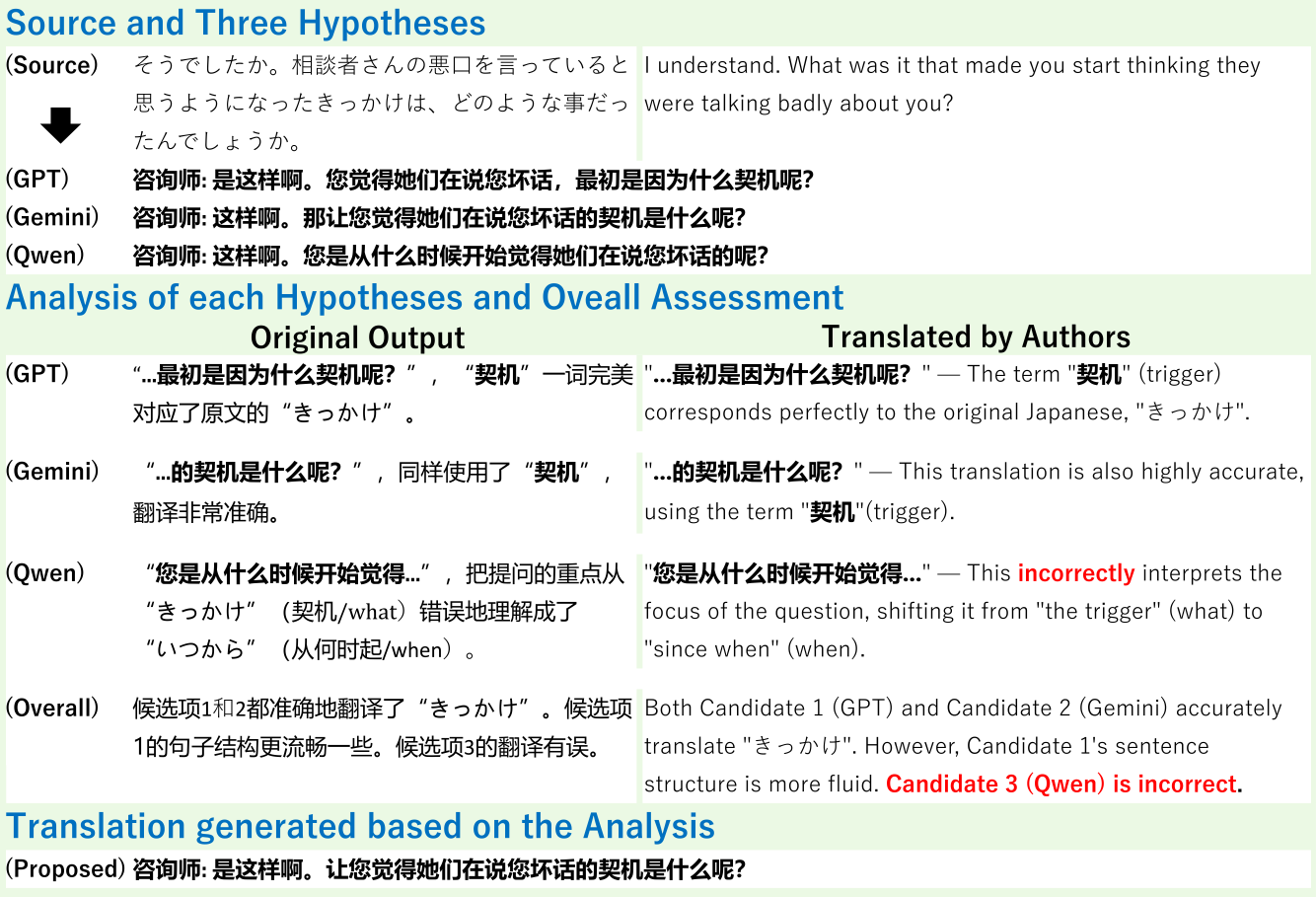}
 \caption{Japanese-to-Chinese translation example 2. This demonstrates the analysis and synthesis process that produced a translation judged inferior by human evaluators to Qwen’s hypothesis due to being slightly unnatural as a conversational text. The refiner LLM rejected Qwen's hypothesis for semantic deviation.}
 \label{case_bad_chinese}
\end{figure*}

\subsection{Human Evaluation}

To assess qualitative aspects that automated metrics cannot capture, such as naturalness and contextual coherence, we conducted a human evaluation using a pairwise comparison framework.
Evaluators were first presented with four turns of the preceding dialogue history. To prevent any particular system's translation style from biasing the context, the dialogue history for each evaluation was constructed using the output of a randomly selected system from the four candidates (proposed method and the three single LLMs). Each evaluator was then presented with a pair of translations: one from proposed method and one from a single LLM.
Without access to the original Japanese text, evaluators were instructed to select the superior translation based on its contextual appropriateness and naturalness within the dialogue history.

The evaluation materials were drawn from 10 randomly selected dialogues from KokoroChat. From these, we sampled 100 unique utterances (10 per dialogue) where the translation from our method was distinct from those of all three single-LLM baselines. Each of these 100 utterances was then systematically paired against the three baseline translations, yielding a total of 300 evaluation pairs per language. This approach ensured that identical translations were never compared.

The assessments were performed by five native Chinese speakers for the Chinese translations. For the English translations, we recruited workers living in English-speaking countries \footnote{Specifically, the United States, Canada, the United Kingdom, Australia, and New Zealand.} through the Amazon Mechanical Turk (MTurk)\footnote{\url{https://www.mturk.com/}}.

\subsection{Results of Human Evaluation}

The results of the pairwise human evaluation, shown in Figures \ref{human_english} and \ref{human_chinese}, were clear and strongly favored the proposed method. For both English and Chinese, the proposed method was strongly and consistently preferred by human evaluators over every single-LLM baseline.
In every head-to-head comparison, the percentage of times the proposed method was judged superior (``Win'') far surpassed the percentage of times it was judged inferior (``Lose''). This powerful result confirms the central hypothesis of the research: the process of integrating and refining outputs from multiple LLMs produces translations that are perceptibly and significantly more natural and contextually appropriate to human readers.

However, it is important to note that some instances where the proposed method was judged inferior were not due to a flaw in its refinement logic, but were a side effect of the randomized experimental design. Figure \ref{case_bad_english} illustrates an example where human evaluators judged the proposed method to be inferior to Grok. From the evaluator's perspective, the counselor appears to substitute the user's negative emotion ("uncool") with a more negative expression ("pathetic"). This apparent error occurs because,to ensure fairness, the dialogue history presented to evaluators was constructed from the output of a randomly selected system. This occasionally created a critical mismatch between the context provided to the evaluator and the internal context upon which our method relied.

\subsection{Case Study}
While the human evaluation results demonstrated the superiority of proposed method, some cases were consistently judged to be inferior. We analyze both successful and unsuccessful cases to explore avenues for future improvement.

Figure \ref{case_good_chinese} illustrates a case where the proposed method functioned as intended, resulting in a translation that was highly preferred by evaluators. The refiner LLM correctly identified the opening of Gemini's translation as the most empathetic among the hypotheses. As noted in the model's internal analysis, this phrase’s use of a modal particle creates a gentle, validating tone, which is more suitable for a counselor than a direct, assertive statement. Furthermore, the final synthesized phrase is more natural and appropriately empathetic than the other candidates. This case demonstrates that the method can successfully analyze the outputs of single LLMs and synthesize a superior utterance suitable for a counselor's role based on that analysis.

Conversely, Figure \ref{case_bad_chinese} presents another instance where the proposed method was judged inferior to a single LLM (Qwen) in Chinese translation. The refiner LLM correctly identifies that Qwen's translation deviates semantically from the source text, shifting the question's focus from ``what'' to ``when.''
Despite this semantic deviation, Qwen's translation was perceived by some human evaluators as a more natural conversational phrase in Chinese. This suggests that the refiner LLM, in this instance, over-prioritized strict semantic fidelity to the source text at the expense of conversational naturalness in the target language. By focusing too heavily on preserving the original question's exact focus, it produced a translation that, while technically more faithful, was ultimately judged inferior to Qwen's more natural-sounding translation.

\section{Discussion}
An analysis of our experimental results reveals a discrepancy between the automated and human evaluations, as well as performance variations across languages. This section examines these findings to discuss the effectiveness of our method.

\subsection{The Limitations of Reference-Free Automated Metrics}

While the proposed method was overwhelmingly preferred in human evaluations, some single-LLM systems achieved comparable or superior scores in the automated evaluation, particularly with the English XCOMET metric. 
This divergence strongly suggests that current reference-free metrics, despite their sophistication, are not yet capable of capturing the full range of qualities that humans perceive as high-quality dialogue.
Specifically, the automatic evaluation metrics used in this study evaluate each utterance in isolation. Consequently, they are ill-equipped to assess factors such as contextual coherence across utterances, appropriateness for speaker roles (counselor or client). Specific examples of these cases are provided in Appendix \ref{appendix2}.
While human evaluation incorporates these aspects by presenting the dialogue history, the automatic approach neglects them entirely. This difference in consideration likely contributes to the divergence between automatic and human evaluation results.

\subsection{The Trade-off Between Semantic Fidelity and Conversational Naturalness}

The translation example in Figure \ref{case_bad_chinese} shows that discrepancies can arise between human evaluators' criteria and the optimization objective of the refiner LLM.
Human evaluators were not shown the source text and based their judgments solely on conversational fluency, so semantic fidelity to the source was not considered. This explains why our proposed method, which is more semantically faithful to the source, was judged inferior in this particular case. 

Conversely, the refiner LLM is provided with the source and prompted to act as a ``professional translator,'' which implicitly biases it toward prioritizing semantic fidelity.
However, the primary purpose of the Multilingual KokoroChat dataset is to serve as a high-quality training corpus for AI counselors in the target languages. For such applications, the dialogue quality in the target language is more important than strict semantic fidelity for our counseling dataset. The example in Figure \ref{case_bad_chinese} suggests that in some cases, sacrificing a degree of fidelity can produce translations that are substantially more natural. A key direction for future work is to develop methods that can better balance this trade-off.

\subsection{Language-Specific Effects}

 The results from both automated and human evaluations indicated that the performance uplift from our ensemble method was more pronounced for Chinese than for English. We hypothesize that this is because the baseline performance of the single LLMs for Japanese-to-English translation is already exceptionally high, leaving less room for improvement. In contrast, for Japanese-to-Chinese, where single-model performance may be slightly less mature, the ensemble's ability to combine strengths and mitigate weaknesses provides a more significant benefit.

This suggests a promising direction for future research: the Multi-LLM Ensemble method may prove to be even more impactful when applied to low-resource language pairs, where the performance of any single translation model is often far from perfect.

\section{Conclusion}

This study introduced a novel, two-stage refinement framework to address the challenge that the optimal LLM varies depending on the input, meaning no single model can consistently guarantee the best possible quality. Our method first generates a diverse set of translation hypotheses using multiple distinct LLMs. Subsequently, a single refiner LLM analyzes the respective strengths and weaknesses of these candidates to synthesize a final, superior translation. By applying this framework to KokoroChat, the world's largest Japanese counseling dialogue corpus, we have created Multilingual KokoroChat, a high-quality, large-scale dataset for English and Chinese.

Our approach’s effectiveness was confirmed through a thorough two-part evaluation. First, automated metrics provided strong quantitative validation, demonstrating that our method was statistically superior or highly competitive. Second, the pairwise human evaluation produced decisive results. Human evaluators overwhelmingly preferred the translations generated by our ensemble method over those from any single LLM, citing their superior naturalness and contextual coherence.

The resulting Multilingual KokoroChat dataset, which is publicly available, is expected to become a valuable resource for the global research community, accelerating advancements in AI-driven mental health support.

\section{Acknowlegments}
I would like to thank Associate Professor Michimasa Inaba of the Department of Informatics, Graduate School of Informatics and Engineering, The University of Electro-Communications, for his continuous guidance and support throughout this research. I am also deeply grateful to all the members of the Inaba Laboratory for their thoughtful suggestions and generous cooperation during the progression of this study. This work was supported by JSPS KAKENHI Grant Number 25H01382.

\section{Bibliographical References}\label{sec:reference}

\bibliographystyle{lrec2026-natbib}
\bibliography{lrec2026-paper}

\section{Language Resource References}
\label{lr:ref}
\bibliographystylelanguageresource{lrec2026-natbib}
\bibliographylanguageresource{languageresource}

\appendix
\section{Appendix}

Section \ref{appendix2} presents examples where translations that are contextually appropriate within a dialogue were assigned low scores by automated metrics because they were evaluated at an utterance-by-utterance level.

\begin{figure*}[t]
 \centering
 \includegraphics[width=0.95\textwidth]{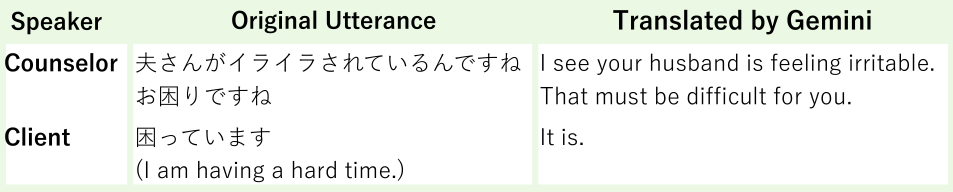}
 \caption{A case where contextually natural translation is penalized by utterance-level evaluation. The translation "It is" captures the conversational flow but be judged as semantically incomplete by reference-free metrics that do not account for the preceding context.}
 \label{auto_eval_bad}
\end{figure*}

\begin{figure*}[t]
 \centering
 \includegraphics[width=0.95\textwidth]{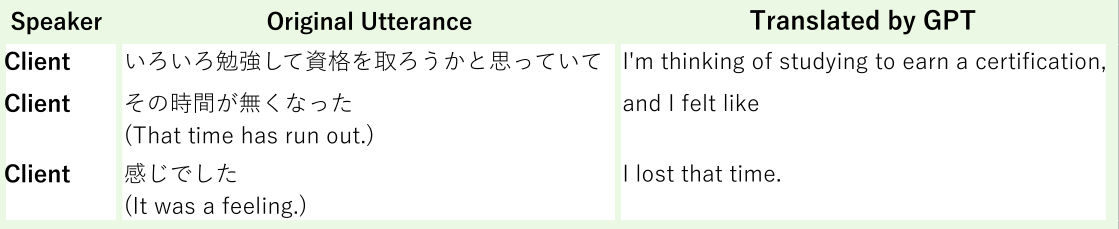}
 \caption{A case where contextually natural translation is penalized by utterance-level evaluation. While the translation elegantly resolves a sentence split, utterance-by-utterance evaluation incorrectly penelize them.}
 \label{auto_eval_bad2}
\end{figure*}

\subsection{Discrepancy between dialogue-level naturalness and utterance-level automatic evaluation.} \label{appendix2}

Figure \ref{auto_eval_bad} and Figure \ref{auto_eval_bad2} illustrate a limitation of reference-free automated metrics that evaluate utterances in isolation. 
In the dialogue shown in Figure \ref{auto_eval_bad}, the Client responds to the Counselor's empathetic remark, "That must be difficult for you." While the original Japanese utterance literally translates to "I am having a hard time," Gemini translates this as "It is." This is a highly natural and contextually appropriate English expression. However, automated metrics evaluating translations on an utterance-by-utterance basis penalize this output for the lack of explicit semantic overlap with the source text. 
Similarly, the dialogue shown in Figure \ref{auto_eval_bad2} demonstrates how speakers naturally divided their utterances into three parts to maintain proper pacing in this text-based counseling format. 
GPT faithfully reproduces this speaker's characteristic utterance segmentation by dividing the entire dialogue into three parts while maintaining the natural speech flow of the conversation. 
However, the second utterance and the third utterance were penalized by automated metrics evaluating in isolation because the translation's meanings no longer matched the individual original Japanese utterance.

These cases highlight a fundamental discrepancy between dialogue-level naturalness and utterance-level automatic evaluation. This discrepancy contributes to the gap between automated evaluations and human evaluation results.

\end{document}